\documentclass{bmvc2k}
\usepackage{tikz}
\usepackage{comment}
\usepackage{amsmath,amssymb} 
\usepackage{color}

\title{Self-distillation and Uncertainty Boosting Self-supervised Monocular Depth Estimation}

\addauthor{Hang Zhou}{hang.zhou@uea.ac.uk}{1}
\addauthor{David Greenwood}{david.greenwood@uea.ac.uk}{1}
\addauthor{Sarah Taylor}{s.l.taylor@uea.ac.uk}{1}
\addauthor{Michal Mackiewicz}{M.Mackiewicz@uea.ac.uk}{1}

\addinstitution{
 School of Computing Sciences
 University of East Anglia\\
 Norwich, UK
}

\runninghead{Zhou et al}{SUB-Depth}

\begin{document}

\maketitle

\begin{abstract}
For self-supervised monocular depth estimation (SDE), recent works have introduced additional learning objectives, for example semantic segmentation, into the training pipeline and have demonstrated improved performance. However, such multi-task learning frameworks require extra ground truth labels, neutralising the biggest advantage of self-supervision. In this paper, we propose \textbf{SUB-Depth}  to overcome these limitations. Our main contribution is that we design an auxiliary self-distillation scheme
and incorporate it into the standard SDE framework, to take advantage of multi-task learning without labelling cost. Then, instead of using a simple weighted sum of the multiple objectives, we employ generative task-dependent uncertainty to weight each task in our proposed training framework.
We present extensive evaluations on KITTI to demonstrate the improvements achieved by training a range of existing networks using the proposed framework, and we achieve state-of-the-art performance on this task.
\end{abstract}

\section{Introduction}
\label{sec:intro}
Depth perception plays an important role in real-world applications including autonomous driving, augmented reality, 3D reconstruction and other high-level computer vision tasks. Although physical sensors such as LiDAR have been deployed widely, estimating depth from pixels is appealing due to the lower cost and compatibility where a camera is available. Supervised depth estimation approaches~\cite{Eigen2014,Miangoleh2021Boosting,chen2021s,GargDualPixelsICCV2019,Ranftl2020,lee2021patch,aich2021bidirectional} can predict dense depth maps but require costly ground truth depth labels. In contrast, self-supervised approaches require no labelled data~\cite{lee2019real,multview,Godard17,tankovich2021hitnet,kendall2017end,zhang2018activestereonet,zhou_sfmlearner,monodepth2,zhou2020constant,zhou_diffnet,lyu2020hr} and are performing competitively.
At a high level, self-supervised depth estimation uses a depth network's output as an intermediate representation for a stereo matching problem or an image reconstruction task. For the latter, the models are trained within a standard self-supervised monocular depth estimation (SDE) framework~\cite{zhou_sfmlearner,monodepth2,zhou2020constant,zhou_diffnet,lyu2020hr,watson2021temporal}. To build such a system, a pose network is introduced to predict the camera pose change between two consecutive frames. Hence, in a standard SDE training framework, a depth network and a pose network are trained simultaneously for \textbf{an image reconstruction} task by optimising the photometric loss.
Previous works~\cite{choi2020safenet,tankovich2021hitnet,liebel2019multidepth,lu_icra} have shown that training a single depth model benefits from multiple regression or classification objectives. Inspired by prior works that train a student depth network with a trained teacher network~\cite{lyu2020hr,poggi2020uncertainty}, we extend the single-task SDE framework to a multi-task setting by introducing a self-distillation scheme associated with a regression objective.
Compared with other multi-task settings which introduce supervised tasks such as semantic segmentation, one of the advantages of self-distillation is that the framework remains a self-supervised regime.

Performance of multi-task systems is dependent on the relative loss weighting for each task. Instead of manually tuning weights of loss terms, inspired by Kendall~\cite{kendall2018multi}, we propose two uncertainty modelling strategies to calculate uncertainty for the self-distillation task and the image reconstruction task respectively. Specifically, the self-distillation uncertainty down-weights the regression loss when a teacher network outputs noisy depth values, and the photometric uncertainty outputs higher confidence where input frames satisfy the image reconstruction tasks' assumptions, that is, static world and ego-motion. We call our system \textbf{SUB-Depth}, and summarise its following key contributions:
\begin{itemize}
 \item We propose a novel multi-task training framework for self-supervised monocular depth estimation.
 \item Instead of manually tuning loss terms' weights, we utilize the task-dependent uncertainty idea, and experiment with several ways of uncertainty modeling.
 \item We conduct exhaustive experiments to show that the proposed training framework is able to boost existing models' performance significantly.
 \end{itemize}

\section{Related literature}
In this section, we review works relating to self-supervised monocular depth estimation, multi-task learning and predictive uncertainty modelling.   
\subsection{Self-supervised monocular depth estimation}
Different from supervised learning based approaches that are trained for a regression task with ground truth depth, self-supervised monocular depth estimation (SDE) methods are trained for an image reconstruction task with a photometric loss. 
Inspired by Structure from Motion (SfM), the seminal work of Zhou~\cite{zhou_sfmlearner} proposed a fundamental framework consisting of a depth network and a pose network which are trained simultaneously with sequential video frames.
Many works have further advanced this framework in several different ways. 
Monodepth2~\cite{monodepth2} introduced minimum-reprojection, structural similarity and multi-scale reconstruction strategies, and is now the most widely used baseline and builds the standard SDE framework.
More recently, a series of works proposed improved network architectures within this framework~\cite{lyu2020hr,guizilini2020,zhou_diffnet}. 
An image distortion handling model was proposed for UnRectDepthNet~\cite{kumar2020unrectdepthnet}, and to guide depth feature learning, Guizilini~\cite{guizilini2020semantically} exploited a semantic segmentation network.
Instead of fixed camera parameters, learnable camera parameters have been used~\cite{chanduri2021camlessmonodepth,gordon2019depth}.
By introducing self-attention and a discrete disparity volume, Johnston and Carneiro further improved Monodepth2~\cite{johnston2020self}.
To boost the single-frame SDE frameworks, Senoh~\cite{multview} proposed a multi-frame training and testing framework. Poggi~\cite{poggi2020uncertainty} introduced uncertainty modelling into SDE approaches, and showed how different strategies impact depth and uncertainty estimation. 

To our best knowledge, we first time propose and incorporate self-distillation into SDE to build a multi-objective learning framework. We demonstrate the performance improvements of our proposed training framework using three existing architectures as underlying models, Monodepth2~\cite{monodepth2}, HR-depth~\cite{lyu2020hr} and DIFFNet~\cite{zhou_diffnet}, which represent base-level, middle-level and high-level performance methods respectively. 
%
\begin{figure*}[t]
\centering
\includegraphics[width=\linewidth]{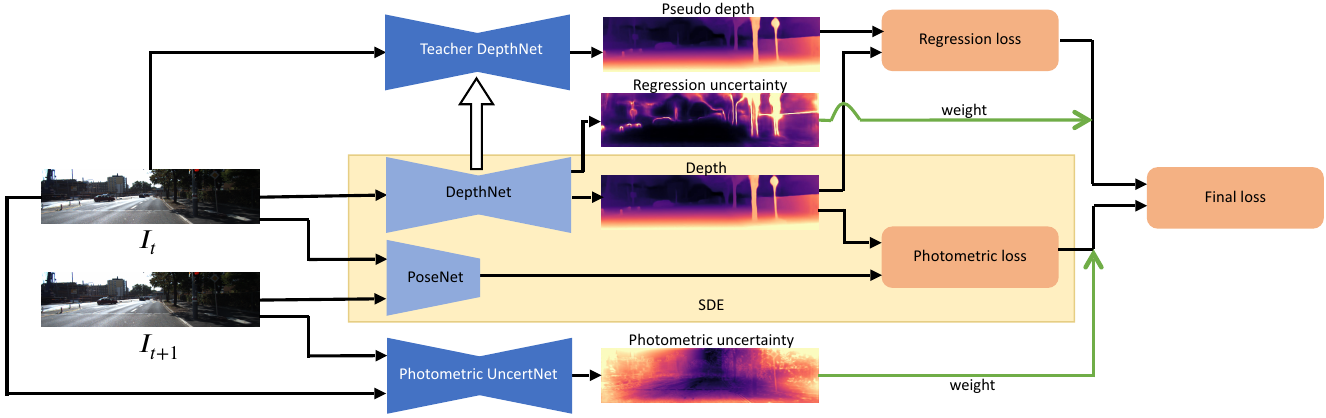}
\caption{An overview of the SUB-Depth framework. SUB-Depth extends the standard existing self-supervised monocular depth estimation framework (SDE) (highlighted) using self-distillation and uncertainty modelling. The teacher DepthNet outputs a supervisory signal for training the DepthNet, and enables computation of a regression loss. Both regression and photometric uncertainty maps are learned and used to weight the respective losses. The teacher DepthNet is pretrained with the highlighted SDE framework by optimising the photometric loss.}
\label{fig:overview}
\end{figure*}
\subsection{Multi-task learning and optimisation}
Visual scene understanding models, trained with multi-task learning systems, always outperform counterparts trained individually~\cite{kendall2018multi}. Recognising similarities with the semantic segmentation task, a series of supervised depth estimation methods~\cite{he2021sosd,kendall2018multi,xu2018pad} and self-supervised methods~\cite{klingner2020self,cai2021x,kumar2021syndistnet,chen2019towards} introduced additional segmentation networks. 
These methods boosted performance by using a shared representation learning encoder for both tasks or by distilling knowledge across tasks. However, building such a multi-task system requires extra semantic annotations, thus losing the most important advantage of self-supervised learning.
Differing from prior works, we exploit knowledge distillation~\cite{hinton2015distilling} as an auxiliary task, in which a student network can learn from a teacher network. In our proposed scheme, as the teacher and student have the same network architecture, we name the additional task \emph{self-distillation}. The teacher network is trained with the standard SDE framework, so this scheme can be implemented without extra labeling cost.

A significant challenge for multi-task learning is how to simultaneously optimise multiple objectives. Sener at el.~\cite{sener2018multi} developed a Frank-Wolfe optimiser to find a Pareto optimal solutions. To minimise the negative conflicts with other gradients during back-propagation, Yu at el.~\cite{yu2020gradient} proposed a form of gradient surgery that projects each task's gradient onto the normal plane of the gradient of any other task and modifies the gradients for each task. In SUB-Depth, we utilise the uncertainty-based weighting approach proposed by Kendall at el.~\cite{kendall2018multi} for jointly learning, which down-weights the task loss contribution where the task-dependent (homoscedastic) uncertainty is high.
\section{SUB-Depth training framework}\label{sec:SUB-depth}
In this section, we first introduce the standard SDE framework, then the proposed self-distillation, and two task-dependent homoscedastic uncertainty formulations. The final system overview is shown in Figure~\ref{fig:overview}.
\subsection{Self-supervised monocular depth estimation}
\label{SDE}
An SDE framework (highlighted by the yellow box in Figure~\ref{fig:overview}) trains a DepthNet $\Theta_{\text{depth}}$ and a PoseNet $\Theta_{\text{pose}}$  simultaneously for an image reconstruction task with a triplet of sequential RGB frames 
\(I_t \in \mathbb{R}^{ H \times W \times 3 },  t \in \{-1, 0, 1\} \). At training time $\Theta_{\text{depth}}$ takes a target frame $I_{0}$ as input and predicts a depth map $d = \Theta_{\text{depth}}(I_{0})$, 
while the PoseNet estimates a relative pose change between the target frame and a source frame, \(T_{0 \rightarrow t^{'}} = \Theta_{\text{pose}}(I_{0},I_{t^{'}}), t^{'} \in \{-1,1\}\). 

Based on the assumption that the world is static and the view change is only caused by a moving camera, a reconstructed counterpart to target frame $I_{0}$ can be generated using only pixels from the source frames \(I_{t^{'}}, t^{'} \in \{-1,1\} \):
\begin{equation}\label{eq:image_warp}
    I_{t^{'} \rightarrow 0 } = I_{t^{'}}[\mathrm{proj}(\mathrm{reproj}(I_{0},d,T_{0 \rightarrow t^{'}}),K)]
\end{equation}
where $K$ are known camera intrinsics, $\left[~\right]$ is the sampling operator, $reproj$ returns a 3D point cloud of camera $t^{'}$, and $proj$ outputs the 2D coordinates when projecting the point cloud onto $I_{t^{'}}$.

Using the predicted depth map $d$, the generated view $I_{t^{'} \rightarrow 0 }$ and the corresponding target frame $I_{0}$, we build a supervisory signal consisting of two items:

\noindent\textbf{Photometric Loss}, $ \ell_{p} $, is an appearance matching loss which calculates the difference between $I_{0}$ and $I_{t^{'} \rightarrow 0 }$. 
Following~\cite{monodepth2,Godard17}, the similarity between a synthesised frame and a target frame is computed using a Structural Similarity term (SSIM)~\cite{wang2004}. Then, combining with the L1 norm, the final photometric loss function is defined: 
\begin{equation}\label{eq:photometrciloss}
\begin{aligned}
    \ell_{p}(I_{0}, I_{t^{'} \rightarrow 0}) &= \alpha \frac{1-\mathrm{SSIM}(I_{0},I_{t^{'} \rightarrow 0})}{2} 
        &+ (1-\alpha)|I_{0} - I_{t^{'} \rightarrow 0}| 
\end{aligned}
\end{equation}
\noindent\textbf{Edge-aware Smoothness}~\cite{Godard17}, $ \ell_{s}$, regularises the depth in low gradient regions:
\begin{equation}\label{eq:smoothness}
    \ell_{s}(d) = |\frac{\nabla d}{\partial x}|e^{-|\frac{\nabla I_{0}}{\partial x}|} + |\frac{\nabla d}{\partial y} |e^{-|\frac{\nabla I_{0}}{\partial y}|}
\end{equation}
We also employ the minimum photometric error, auto-masking and multi-scale depth loss techniques which were introduced in~\cite{monodepth2}. The final self-supervised photometric objective is defined:
\begin{equation}\label{eq:lossfunction}
    \begin{aligned}
    \ell_{photometric} &= \mathrm{min}( \ell_{p}(I_{0}, I_{t^{'} \rightarrow 0}) ) 
        &+ \beta\ell_{s}(d), t^{'} \in \{-1,1\}
    \end{aligned}
\end{equation}
where $\beta$ is a weighting coefficient between the photometric loss $\ell_{p}$ and depth smoothness $\ell_{s}$. The objective loss is averaged per pixel, pyramid scale and image batch.
\subsection{Self-distillation scheme}
\label{self-distillation}
Most related works focus on integrating other supervised learning based tasks into an SDE framework. 
Typically, when introducing a segmentation task, the segmentation network shares the encoder in the SDE depth network, and
all components are trained jointly with the sum of the photometric loss and the segmentation loss. 
Although depth models trained with such a multi-task system can improve their performance, it neutralises the advantage of SDE frameworks.

Unlike existing multi-task strategies, self-distillation avoids introducing extra manual annotations. 
Instead, we use an SDE trained teacher depth network $\Theta_{\text{teacher}}$ to output pseudo depth ground truth $d_{pseudo} = \Theta_{\text{teacher}}(I_{0})$. 
We then let the depth map from the DepthNet $d = \Theta_{\text{depth}}(I_{0})$ regress the $d_{pseudo}$. The objective can be formulated as an L1 \textbf{regression loss}:
\begin{equation}\label{eq:regression_loss}
    \ell_{regression} = |\Theta_{\text{depth}}(I_{0}) - \Theta_{\text{teacher}}(I_{0})|
\end{equation}
As $\Theta_{\text{teacher}}$ and $\Theta_{\text{depth}}$ have the same network architecture, we name this task \textbf{self-distillation}. 

By simply introducing a $\Theta_{\text{teacher}}$, we retrain depth networks using following weighted loss function:
\begin{equation}\label{eq:weighted_loss}
    \ell = \omega_{pho} \times \ell_{photometric} + \omega_{reg} \times \ell_{regression}
\end{equation}
Where $\omega_{pho}$ and $\omega_{reg}$ are weights for $\ell_{photometric}$ and $\ell_{regression}$ respectively. We train and evaluate models using different weighting settings, shown in Table~\ref{tab:mannually_tune}. From the table, we observe that this naive multi-task learning framework can output a $\Theta_{\text{depth}}$ which outperforms the $\Theta_{\text{teacher}}$ trained with standard SDE framework, no matter what the ratio is. However, when we set $\omega_{pho} = 0.2$ and $\omega_{reg} = 0.8$, models gain best performance for Rel Abs, while they are improved significantly for $\delta_{1}$ when $\omega_{pho} = 0.6$ and $\omega_{reg} = 0.4$. As it is hard to get an optimal weight settings, we utilize uncertainty based methods to automatically balance loss terms.
%

\begin{table*}[tb]
\caption{Comparison between manually tuned objective weights, evaluated on the KITTI \cite{kitti} Eigen split. We experiment with several combinations of $\omega_{pho}$ and $\omega_{reg}$. The best weighting pairs are in \textcolor{red}{red}. The best (Rel Abs and $\delta_{1}$) scores are \underline{\textbf{bold and underlined}}. Error and accuracy metrics' definitions are given in \ref{sec:metrcis}.}
\label{tab:mannually_tune}
\centering
\setlength
\tabcolsep{4.2pt}{
\begin{tabular}{c c|c c c c |c c  c}
\hline
 \multicolumn{2}{c|}{Objective weights}& \multicolumn{4}{c|}{Error metrics}& \multicolumn{3}{c}{Accuracy metrics}\\
$\omega_{pho}$ & $\omega_{reg}$& Rel Abs & Sq Rel&  RMSE & RMSE log & $\delta_{1}$ & $\delta_{2}$ & $\delta_{3} $  \\\hline
0 & 1 &   0.112  &   0.884  &   4.740  &   0.189  &   0.881  &   0.961  &   0.982  \\
\textcolor{red}{0.2} & \textcolor{red}{0.8} &   \underline{\textbf{0.110}}  &   0.855  &   4.724  &   0.188  &   0.881  &   0.961  &   0.982  \\
0.4 & 0.6 &   0.112  &   0.866  &   4.736  &   0.189  &   0.881  &   0.961  &   0.982  \\
0.5 & 0.5 &   0.112  &   0.888  &   4.766  &   0.189  &   0.882  &   0.961  &   0.981  \\
\textcolor{red}{0.6} & \textcolor{red}{0.4} &  0.113  &   0.876  &   4.774  &   0.189  &  \underline{\textbf{0.884}}  &   0.962  &   0.983  \\
0.8 & 0.2 &   0.113  &   0.885  &   4.799  &   0.190  &   0.882  &   0.961  &   0.981  \\
1 & 0 &   0.115  &   0.903  &   4.863  &   0.193  &   0.877  &   0.959  &   0.981\\\hline
\end{tabular}}
\end{table*}

\subsection{Task-dependent uncertainty formulation} \label{sec:uncertainty}
Following~\cite{kendall2017uncertainties}, given a dataset $(x,y)$, 
we let the network output the mean $\hat{y}$ and the variance $\sigma$ of a posterior probability distribution $p(y|\hat{y},\sigma)$ over ground truth $y$, which can be modelled as Laplacian or Gaussian. 
If Laplace's distribution:
\begin{equation}\label{eq:likelihood}
    p(y|\hat{y},\sigma) = \frac{1}{2\sigma}\mathrm{exp}\frac{-|\hat{y} - y|}{\sigma}
\end{equation}
is used, then the network can be trained by minimising the loss~\cite{klodt18}:
\begin{equation}\label{eq:loss}
    loss = \frac{|\hat{y} - y|}{\sigma} + \mathrm{log}(\sigma)
\end{equation}
where the variance $\sigma$ increases when the ground truth $y$ is unreliable. 
As a result, we can treat $\sigma$ as task-dependent uncertainty, and the penalty term $\mathrm{log}(\sigma)$, avoids the degenerate solution $\sigma = + \infty$.

We introduce uncertainty modelling for each sub-task in the framework:

\noindent\textbf{Uncertainty for image reconstruction}. Intuitively, as photometric loss is a measurement of the difference between two images, it is natural to estimate its uncertainty with a model that takes two images as input. While prior works~\cite{poggi2020uncertainty,yang2020d3vo} use the DepthNet $\Theta_{\text{depth}}$ for modelling the photometric uncertainty, we propose a separate Photometric UncertNet $\Theta_{\text{pho}}$ to estimate the uncertainty. As for the input of the proposed uncertainty network, we experiment with different settings: 1). feeding the target frame $I _{t}$, 2). feeding the target $I_{t}$ and aligned $I_{t'}$ (see in supplementary material for more details). Finally, we let UncertNet take the target frame and the source frame as inputs and output the photometric uncertainty map $\sigma_{pho}$, as shown in Figure~\ref{fig:overview}. 
Then the uncertainty weighted photometric loss, with the penalty term $\mathrm{log}(\sigma_{pho})$, for the image reconstruction task is given by:
\begin{equation}\label{eq:uncerted_photometric}
    \ell_{reconstruction} = \frac{\ell_{photometric}}{\sigma_{pho}} + \log(\sigma_{pho})
\end{equation}
\noindent\textbf{Uncertainty for self-distillation}. 
We let the DepthNet $\Theta_{\text{depth}}$ encode and output depth regression uncertainty $\sigma_{reg}$. Besides, we explore using a standalone regression uncertainty network to estimate depth uncertainty (see in supplementary material for more details). Then the uncertainty weighted regression loss with the penalty term $\mathrm{log}(\sigma_{reg})$ for the self-distillation task can be computed as:   
\begin{equation}\label{eq:uncerted_regress}
   \ell_{distillation} = \frac{\ell_{regression}}{\sigma_{reg}} + \mathrm{log}(\sigma_{reg})
\end{equation}
\subsection{Multi-task learning with uncertainty}
Finally we combine the uncertainty weighted photometric loss ($\ell_{reconstruction}$) and regression loss ($\ell_{distillation}$) to build \textbf{SUB-Depth}:
\begin{equation}\label{eq:final_loss}
   \ell_{final} = \ell_{reconstruction}  + \ell_{distillation}
\end{equation}
The result is a multi-task learning system, which trains $\Theta_{\text{depth}}$ for an image reconstruction task and a self-distillation task using the sum of task-dependent uncertainty weighted losses.

\begin{figure*}[tb]
\begin{center}
\includegraphics[trim=28 0 13 0, clip, width=.49\columnwidth]{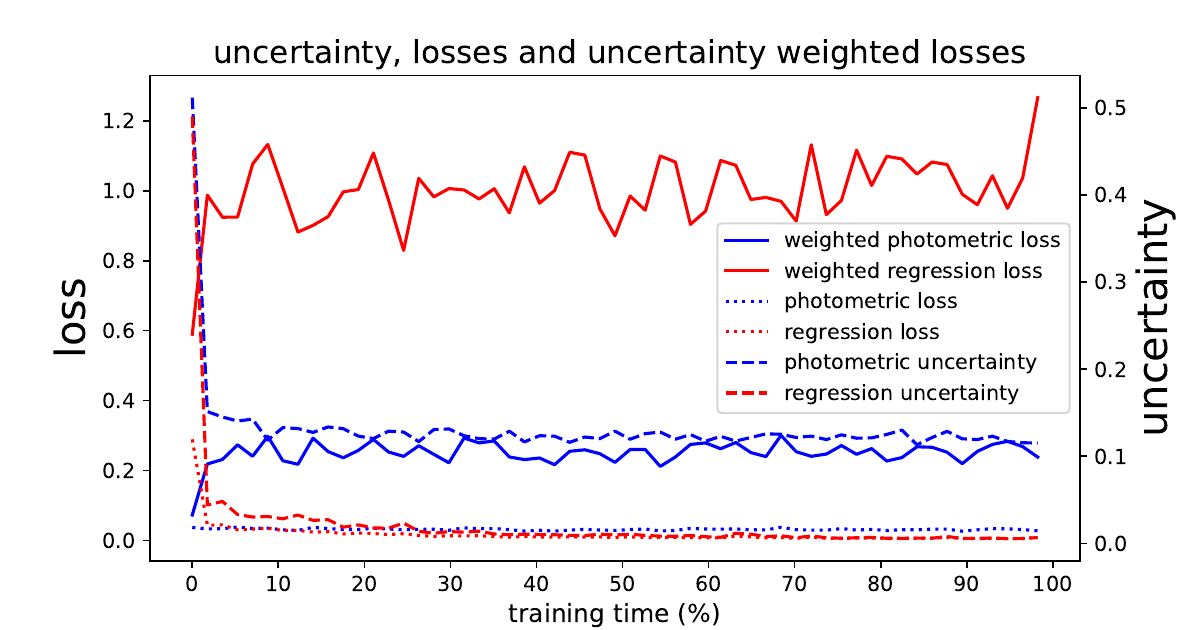}
\hfill
\includegraphics[trim=28 0 30 0, clip,
width=.49\columnwidth]{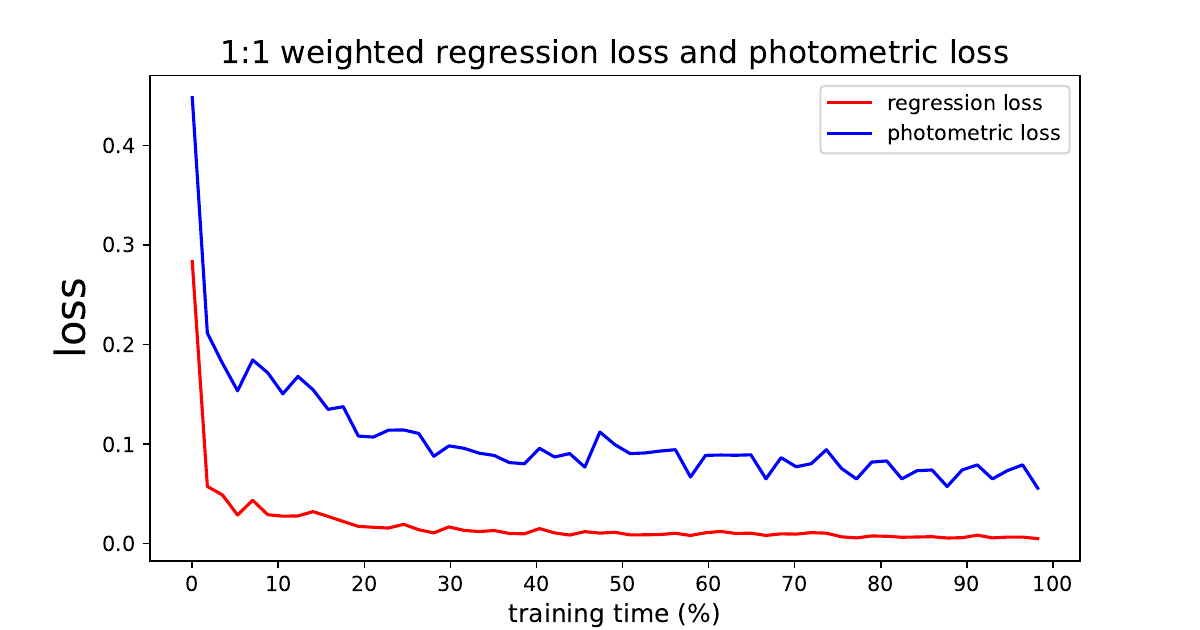}
\end{center}
\caption{
Left: The task-dependent losses, uncertainty weighted losses and uncertainty estimates during SUB-Depth training. Right: The corresponding task-dependent losses of the same system trained with no uncertainty modelling. Uncertainty modelling increases the contribution of the regression loss, and down-weights photometric loss.
}
\label{fig:loss_uncertainty}
\end{figure*}

The difference during training between the naive unweighted sum of losses and uncertainty weighted losses is shown in Figure~\ref{fig:loss_uncertainty}.
On the left plot, $\Theta_{\text{depth}}$ is trained with self-distillation as a prime task. 
In this graph we observe that, although the unweighted regression loss is lower than the unweighted photometric loss throughout most of the training, after applying the task-dependent uncertainty weighting the self-distillation task contributes more to the $\ell_{final}$ than the reconstruction loss. 
This change is due to the regression uncertainty $\sigma_{reg}$ being lower than the photometric uncertainty $\sigma_{pho}$, and indicates that pseudo-labels from the teacher DepthNet provide a more reliable supervisory signal than the pixel-level metrics used in the photometric loss.
For comparison,  the  right  plot  in  Figure~\ref{fig:loss_uncertainty} shows the naive 1:1 weighted multi-task training framework without uncertainty modelling.
In this case, the photometric task dominates the loss throughout training following similar curves to the respective unweighted losses on the left plot. 
\section{Implementation}\label{sec:implement}
Our models are trained and tested on a single NVIDIA RTX 6000 GPU using PyTorch~\cite{paszke2019pytorch}. 
A depth network and a pose network are trained for 20 epochs using the Adam optimiser~\cite{kingma2015} with the default betas $0.9$ and $0.999$. 
They were trained with a batch size of 8 and an input and output resolution of \( 640 \times 192 \).
We set the initial learning rate as \( 10^{-4} \) for the first 14 epochs and then \( 10^{-5} \) for fine-tuning the remainder.
In the objective function $\ell_{final}$ (Equation~\ref{eq:final_loss}), we set the SSIM weight $\alpha = 0.85$ and the edge-aware smoothness weight $\beta = 1\times 10^{-3}$.

\noindent\textbf{DepthNet and Teacher DepthNet}. To verify the generalisation capability of SUB-Depth, we train three different architectures: Monodepth2~\cite{monodepth2}, HR-depth~\cite{lyu2020hr} and DIFFNet~\cite{zhou_diffnet}, which represent baseline-level, mid-level and state-of-the-art methods when trained with the standard SDE framework. DepthNet models are initialised on the Imagenet~\cite{imagenet} pretrained weights. The teacher DepthNets are fixed models that are pretrained with the SDE framework. To generate the associated regression uncertainty, we modify output layers which originally produce one-channel depth maps to two-channel output.

\noindent\textbf{PoseNet and Photometric UncertNet}. For all training settings, we implement the architecture proposed in~\cite{monodepth2} for pose estimation, which is built on ResNet-18. The pose network takes the two adjacent frames as input and outputs the relative pose which is parameterised with a 6-DOF vector. The photometric uncertainty network uses an encoder-decoder with skip-connections. The encoder is based on the ResNet-18 architecture and the decoder follows the design of the Monodepth2 depthnet decoder~\cite{monodepth2}. The photometric uncertainty network takes adjacent frames as input and outputs photometric uncertainty maps.
\section{Experiments and results}
In this section we describe and evaluate our framework on the KITTI dataset. We explore the observed improvements in performance, and perform an ablation study to determine the contribution of each component of the SUB-Depth training framework.
\subsection{Dataset and metrics}\label{sec:metrcis}
{\bf KITTI}~\cite{kitti} is a dataset that contains stereo images and corresponding 3D laser scans of outdoor scenes captured by imaging equipment mounted on a moving vehicle~\cite{kingma2015}. The RGB images have a resolution of $\approx 1241 \times 376$ and the corresponding depth maps are sparse with a large amount of missing data. For training, we adopt the dataset split proposed by~\cite{Eigen2014} and resize images to $ 640 \times 192$. After removing the static frames by a pre-processing step suggested by~\cite{zhou_sfmlearner}, this results in 39,810 monocular frame triplets for training and 4,424 frame triplets for validation. To simplify the training process, the camera intrinsic matrices are assumed identical for all the frames in different scenes. To obtain this ``universal'' intrinsic matrix, we offset the principal point of the camera to the image centre and reset the focal length as the average of all the focal lengths in KITTI. 

\textbf{Depth metrics} described by Eigen~\cite{Eigen2014} are the most common used metrics for evaluating depth estimation accuracy. They include four error metrics: the Absolute Relative Error (Abs Rel), Squared Relative Error (Sq Rel), Root Mean Squared Error (RMSE), and the log of RMSE; accuracy metric: $\delta_{1}$, $\delta_{2}$, $\delta_{3}$. We report each of these measures for each setting in our evaluation.

\textbf{Uncertainty metric}. Although uncertainty modelling is not our main contribution, we validate and compare the uncertainty outputs with two selected methods from Poggi et al.~\cite{poggi2020uncertainty}. When evaluating uncertainty, we treat the depth regression uncertainty from DepthNet $\Theta_{\text{depth}}$ as depth uncertainty. From Ilg et al.~\cite{Ilg_2018_ECCV}, we use the Area Under the Sparsification Error (AUSE), the lower the better, and the Area Under the Random Gain (AURG), the higher the better, to quantify the uncertainty modelling performance of three depth metrics: Abs Rel, RMSE and $\delta_{1}$, respectively in Table~\ref{tab:uncertainty}.
%
%
%
\subsection{Evaluation on KITTI}
To evaluate the performance of SUB-Depth, we select and retrain three model architectures from prior work using our training framework: Monodepth2~\cite{monodepth2}, HR-depth~\cite{lyu2020hr} and DIFFNet~\cite{zhou_diffnet}. 
In each case, when compared to the original model (teacher DepthNet), we see significant improvements in all metrics. Table~\ref{tab:quantity} displays this quantitative comparison for all standard metrics for KITTI. We particularly draw attention to the improvement for DIFFNet, a recent state-of-the-art model, that still exhibits substantial improvement. DIFFNet trained using SUB-Depth establishes a new level of performance on the KITTI corpus.
\begin{table*}[tb]
\caption{\textbf{Quantitative comparison of SUB-Depth to existing SDE framework trained models on KITTI~\cite{kitti} Eigen split}. The best results in each subsection are in \textbf{bold}. Models trained with SUB-Depth outperform the same models trained with SDE in every case.} 
\label{tab:quantity}
\centering
\setlength
\tabcolsep{4.2pt}{
\begin{tabular}{l| c c c c |c c c}
\hline
 Method  &Abs Rel& Sq Rel&  RMSE & RMSE log & $\delta_{1}$ & $\delta_{2}$ & $\delta_{3} $  \\
\hline
Monodepth2~\cite{monodepth2} & 0.115 & 0.903 & 4.863 & 0.193 & 0.877 & 0.959 & 0.981\\
+ SUB-Depth &    \textbf{0.110} &   \textbf{0.821}  &   \textbf{4.648}  &   \textbf{0.185}  &   \textbf{0.884}  &  \textbf{0.962}  &   \textbf{0.983}   \\
\hline
Improvement & 0.005 & 0.082 & 0.115 & 0.008 & 0.007 &0.003 & 0.002 \\ \hline \hline
HR-depth~\cite{lyu2020hr} & 0.109 & 0.792 & 4.632 & 0.185 & 0.884 & 0.962 & 0.983\\
+ SUB-Depth &    \textbf{0.106}  &   \textbf{0.770}  &   \textbf{4.545}  &   \textbf{0.182}  &   \textbf{0.888}  &   \textbf{0.963}  &   \textbf{0.983}  \\\hline
Improvement & 0.003 & 0.022 & 0.087 & 0.003 & 0.004 &0.001 & 0 \\ \hline \hline
DIFFNet~\cite{zhou_diffnet}&   0.102  &   0.764  &   4.483  &   0.180  &   0.896  &   0.965  &  0.983  \\
+ SUB-Depth  &  \textbf{0.099}  &   \textbf{0.695}  &   \textbf{4.326}  &   \textbf{0.175}  &   \textbf{0.900}  &   \textbf{0.966}  &   \textbf{0.984}  \\\hline
Improvement& 0.003 & 0.059 & 0.157 & 0.005 & 0.004 &0.001 & 0.001 \\ 
\hline
\end{tabular}}
\end{table*}
In Table~\ref{tab:uncertainty}, We evaluate the uncertainty modelling performance on three different depth metrics. With respect to AUSE, our proposed method outperforms other competitors from Poggi et al.~\cite{poggi2020uncertainty}, while, for AURG, there is a marginal gap between ours and the Self method.

\begin{table*}[tb]
\caption{\textbf{Quantitative comparison of uncertainty modelling.} We evaluate two uncertainty metrics for each selected depth metric and compare with two uncertainty modelling methods (Log and Self) in~\cite{poggi2020uncertainty}. AUSE is lower the better, and AURG is higher the better.}
\label{tab:uncertainty}
\centering
\setlength
\tabcolsep{2pt}{
\begin{tabular}{l|c c|c c|c c}
\hline
~ & \multicolumn{2}{c|}{Abs Rel} & \multicolumn{2}{c|}{RMSE} & \multicolumn{2}{c}{$\delta_{1}$} \\
\hline
Method & AUSE & AURG & AUSE & AURG & AUSE & AURG\\\hline
Poggi-Log~\cite{poggi2020uncertainty} & 0.051 & 0.027 & 3.097 & 1.188 & 0.060 & 0.056\\
Poggi-Self~\cite{poggi2020uncertainty} & 0.036 & 0.038 & 2.292 & 1.779 & 0.037 & 0.072\\
\hline\hline
SUB-Depth &   0.035  &   0.037  &   2.196  &   1.770  &   0.034  &   0.072  \\
\hline

\end{tabular}}
\end{table*}
\begin{table*}[tb]
\caption{\textbf{Quantitative comparison of uncertainty modelling on improved ground truth~\cite{uhrig2017sparsity}}.}
\label{tab:uncertainty_improved}
\centering
\setlength
\tabcolsep{2pt}{
\begin{tabular}{l|c c|c c|c c}
\hline
~ & \multicolumn{2}{c|}{Abs Rel} & \multicolumn{2}{c|}{RMSE} & \multicolumn{2}{c}{$\delta_{1}$} \\
\hline
Method & AUSE & AURG & AUSE & AURG & AUSE & AURG\\\hline

Poggi-Log~\cite{poggi2020uncertainty}  & 0.039 & 0.020 & 2.562 & 0.916 & 0.044 & 0.038 \\
Poggi-Self~\cite{poggi2020uncertainty} & 0.030 & 0.026 & 2.009 & 1.266 & 0.030 & 0.045\\
\hline\hline
SUB-Depth &  0.029  &   0.026  &   1.950  &   1.245  &   0.028  &   0.045  \\
\hline

\end{tabular}}
\end{table*}
\begin{table*}[htb!]
\caption{\textbf{Ablation Studies}. We observe increased performance as self-distillation is introduced, and further improvements with the addition of uncertainty modelling. The best results in each subsection are
in \textbf{bold}.}
\label{tab:ablation}
\centering
\setlength
\tabcolsep{2pt}{
\begin{tabular}{l| c  c c c |c c  c}
\hline
Objective &Abs Rel& Sq Rel&  RMSE & RMSE log & $\delta_{1}$ & $\delta_{2}$ & $\delta_{3} $  \\
\hline
$\ell_{photometric}$(Baseline)&   0.115  &   0.903  &   4.863  &   0.193  &   0.877  &   0.959  &   0.981\\
$\ell_{regression}$  &   0.112  &   0.884  &   4.740  &   0.189  &   0.881  &   0.961  &   0.982  \\\hline
Ours(1:1 weighted) &  0.112  &   0.888  &   4.766  &   0.189  &   0.882  &   0.961  &   0.981  \\
Ours(uncertainty weighted) & \textbf{0.110} &   \textbf{0.821}  &   \textbf{4.648}  &   \textbf{0.185}  &   \textbf{0.884}  &  \textbf{0.962}  &   \textbf{0.983}  \\\hline
\end{tabular}}
\end{table*}

To validate the performance improvements gained by SUB-Depth and evaluate the contribution of each design, we conduct an ablation study as shown in Table~\ref{tab:ablation}.
Monodepth2~\cite{monodepth2} is used as the underlying architecture for all results reported in this table.
The first row $\ell_{photometric}$ is the result from the standard SDE framework, and performs the worst of all settings.
In second row $\ell_{regression}$, by simply using the trained DepthNet as a teacher DepthNet we achieve improved performance across all measures.
In last two rows, performance improves further as $\ell_{photometric}$ and $\ell_{regression}$ are combined and weighted by corresponding uncertainty estimation.

We offer additional evaluation on top-10 challenging subset~\cite{zhou_diffnet} of KITTI, qualitative results on KITTI, generalisation results on Cityscapes~\cite{cordts2016cityscapes} and visualisation of error maps on Virtual KITTI~\cite{Gaidon:Virtual:CVPR2016} that are reported in supplementary material.
%

\section{Conclusion}
We presented a multi-task training framework for self-supervised monocular depth estimation, SUB-Depth. SUB-Depth extends the existing standard depth estimation framework with the introduction of self-distillation and uncertainty modelling. We introduce a teacher network and let the depth network be trained, not only for an image reconstruction task but also for a self-distillation task. To find the optimal objective weights, we utilize task-dependent uncertainty to weight losses for each task. Through analysing losses and uncertainty during training, we discovered that, initially the image reconstruction task contributes more than the self-distillation task, but then self-distillation quickly becomes the primary task since the estimated regression uncertainty is much lower than photometric uncertainty.
We retrained three representative approaches using SUB-Depth to validate the generalisation capability of our proposed framework, and all outperform their counterparts.
Our SUB-Depth training framework exhibits substantial improvements over the current state-of-the-art model on the KITTI benchmark for all depth metrics.

\textbf{Acknowledgement}\\
The research presented in this paper was carried out on the High Performance Computing Cluster supported by the Research and Specialist Computing Support service at the University of East Anglia. .

\bibliography{BMVC2022}
\end{document}


\maketitle
In Table \textcolor{red}{1}, in addition to our final uncertainty modelling scheme (last row), we experiment with two input settings for proposed Photometric UncertNet (first two rows) and a standalone uncertainty network (3rd row) for self-distillation uncertainty modelling.

\begin{table*}[tbh]
\centering
\label{tab_settings}
\caption{SUB-Depth experiments with different uncertainty settings. 1: feeding first frame into UncertNet. 2: feeding first frame and aligned frame into UncerNet. 3: standalone regression uncertainty network.}
\setlength
\tabcolsep{3.5pt}{
\small
\begin{tabular}{l|c  c c  |c c  c}
\hline
idx &Abs Rel& Sq Rel& RMSE log &  $\delta_{1}$ &  $\delta_{2}$ & $\delta_{3} $  \\
\hline
1 &   0.113  &   0.905   &   0.189  &   0.882  &   0.961  &   0.982  \\
2 &    0.111  &   0.875  &   0.188  &   0.882  &   0.960  &   0.982    \\
3 &   0.111  &   0.870  &  0.188  &   0.883  &   0.961  &   0.982  \\
\hline
Final & \textbf{0.110} &   \textbf{0.821}  &   \textbf{0.185}  &   \textbf{0.884}  &  \textbf{0.962}  &   \textbf{0.983}  \\\hline
\end{tabular}}
\end{table*}

In Table~\ref{tab:comparison_hard}, we extended our quantitative evaluation by selecting the top 10 most challenging images for each model, following the method described by ~\cite{zhou_diffnet}. The top 10 hardest images show areas of deep shadow, poor lighting, foliage and other photographically indistinct regions. Our method deals with this uncertainty and improves on the results of all prior methods for this subset of the benchmark test set.
\begin{table*}[tbh]
\caption{\textbf{Quantitative comparison of SUB-Depth to existing SDE framework trained models on top-10 selected subset of KITTI~\cite{kitti} benchmark}. The best results in each subsection are in \textbf{bold}. Models trained with SUB-Depth outperform the same models trained with SDE in every case.}
\label{tab:comparison_hard}
\centering
\setlength
\tabcolsep{4.2pt}{
\begin{tabular}{l|c c c c |c c  c}
\hline
 Method  &Abs Rel& Sq Rel&  RMSE & RMSE log & $\delta_{1}$ & $\delta_{2}$ & $\delta_{3} $  \\\hline
Monodepth2~\cite{monodepth2} &   0.250  &   3.008  &   7.515  &   0.353  &   0.683  &   0.870  &   0.924  \\
+ SUB-Depth  & \textbf{0.229}  &   \textbf{2.451}  &   \textbf{6.885}  &   \textbf{0.330}  &   \textbf{0.713}  &   \textbf{0.876}  &   \textbf{0.931}    \\\hline
Improvement & 0.021 & 0.557 & 0.63 & 0.023 & 0.030 & 0.006 & 0.007 \\ \hline \hline
HR-depth~\cite{lyu2020hr}&   0.240  &   1.687  &   5.433  &   0.320  &   0.669  &   0.871  &   0.947  \\
+ SUB-Depth &   \textbf{0.222}  &   \textbf{1.566}  &   \textbf{5.176}  &   \textbf{0.304}  &   \textbf{0.710}  &   \textbf{0.891}  &   \textbf{0.949}  \\\hline
Improvement & 0.018 & 0.121 & 0.257 & 0.016 & 0.041 & 0.020 & 0.002\\ \hline \hline
DIFFNet~\cite{zhou_diffnet} &   0.225  &   2.160  &   6.357  &   0.312  &   0.712  &   0.899  &   0.951   \\
+ SUB-Depth &    \textbf{0.209}  &   \textbf{1.672}  &   \textbf{5.783}  &   \textbf{0.294}  &   \textbf{0.723}  &   \textbf{0.907}  &   \textbf{0.957}    \\\hline
Improvement & 0.016 & 0.488 & 0.574 & 0.018 & 0.011 & 0.008 & 0.006 \\ \hline 
\end{tabular}}
\end{table*}

\begin{figure*}[tb]
\centering
\includegraphics[width=\linewidth]{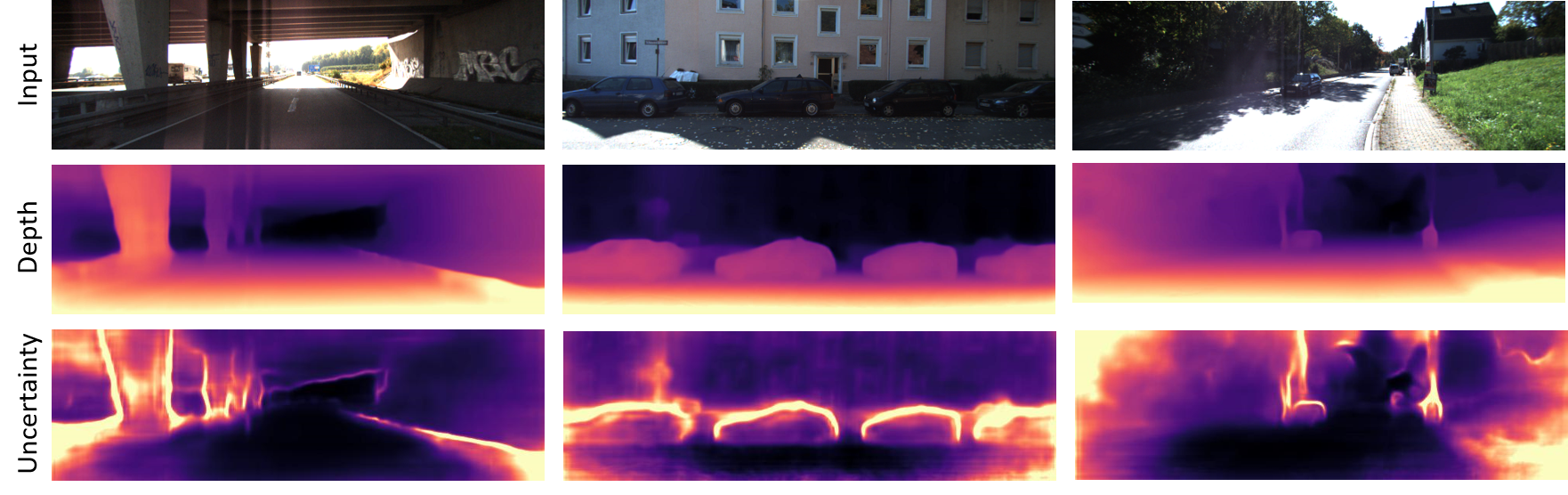}
\caption{\textbf{Qualitative results on KITTI~\cite{kitti}}. We visualise the depth and the uncertainty maps from SUB-Depth trained Monodepth2. The uncertainty maps capture high uncertainty at object boundaries with a hotter color.}
\label{fig:quality}
\end{figure*}

\begin{figure*}[tb]
\centering
\includegraphics[width=\linewidth]{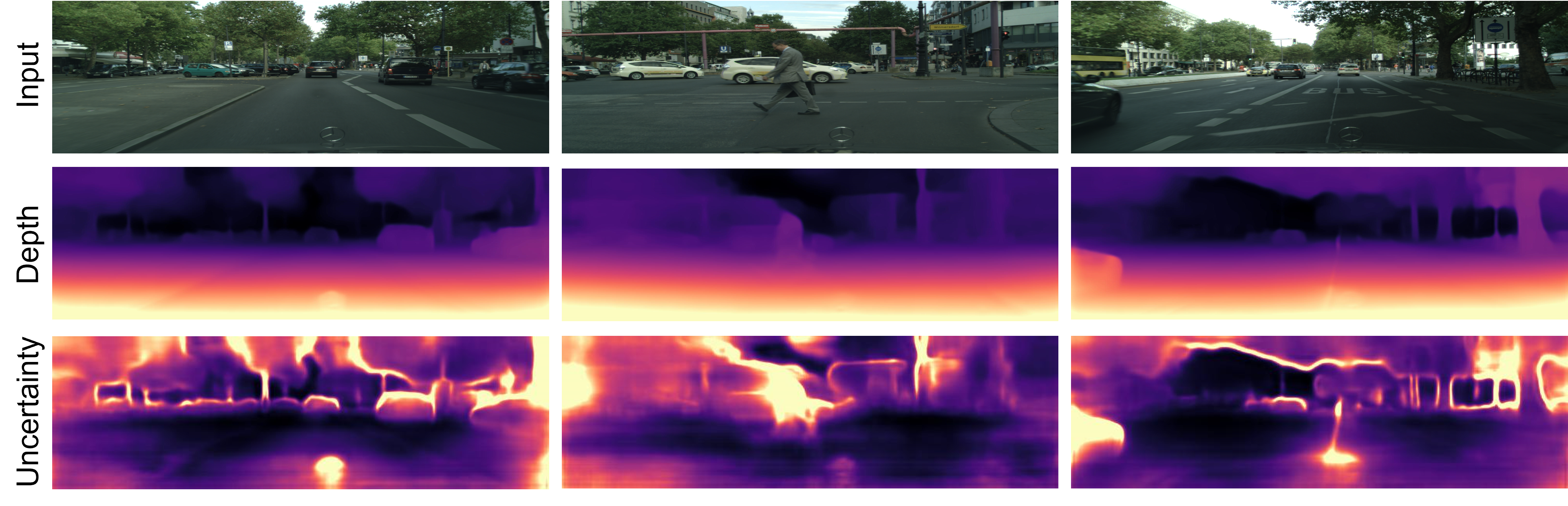}
\caption{\textbf{Generalisation results on Cityscapes~\cite{cordts2016cityscapes}}. We visualise the depth and the uncertainty maps from SUB-Depth trained only with KITTI. The uncertainty maps show higher
uncertainty with a hotter color, and illustrate greater uncertainty at object boundaries and for moving objects.}
\label{fig:cs_result}
\end{figure*}

Qualitative evaluations are provided in Figure~\ref{fig:quality} for randomly selected examples. For each example, we show input RGB and output depth and regression uncertainty maps. The uncertainty map correctly marks object boundaries with high values where the transition from near to far distance is more difficult to predict. To show generalisation performance, in Figure~\ref{fig:cs_result}, we additionally qualitatively evaluate the same depth network on the Cityscapes dataset~\cite{cordts2016cityscapes}. Although trained only on KITTI, the model appears to generalise well for both depth estimation and uncertainty modelling.

\begin{figure*}[tb]
\centering
\includegraphics[width=\linewidth]{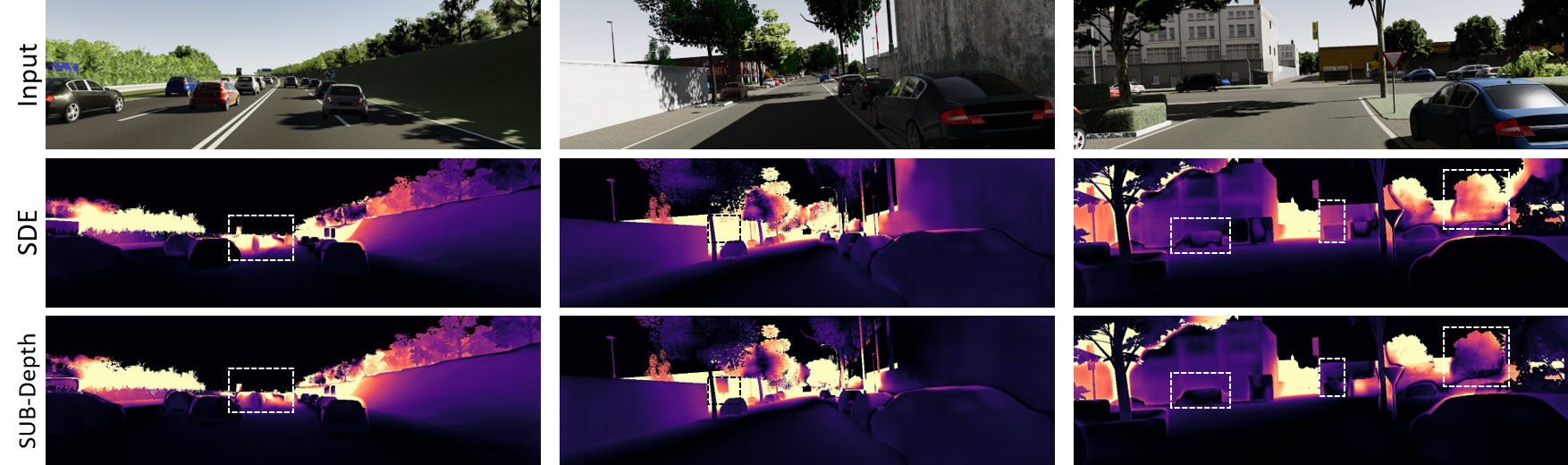}
\caption{\textbf{Visualisation of error map on Virtual KITTI~\cite{Gaidon:Virtual:CVPR2016}}. The top row contains the synthetic input images. The second row shows the Abs rel error maps from SDE trained Monodepth2. The bottom row shows the error maps from SUB-Depth trained Monodepth2. The differences are highlighted by white dotted boxes.}
\label{fig:err_map_comparison}
\end{figure*}

As KITTI does not have dense ground truth depth maps, we use Virtual KITTI~\cite{Gaidon:Virtual:CVPR2016} to compute depth error maps in Figure~\ref{fig:err_map_comparison}. In this qualitative evaluation we show, from top to bottom, the input RGB image, the depth error maps from the baseline Monodepth2 model and the error maps from Monodepth2 trained with SUB-Depth. For each randomly selected example, we highlight regions of the depth maps that show compelling improvements over prior work. The images are provided at high resolution to allow the reader to zoom in.
\bibliography{BMVC2022}